\documentclass[a4paper, 10pt, conference]{ieeeconf}      
\usepackage{FG2023}

\usepackage[small]{caption}
\usepackage{graphicx}
\usepackage{amsmath}
\usepackage{booktabs}
\usepackage{amssymb}
\usepackage{amsmath}
\usepackage{amsfonts}
\usepackage{bm} 

\def\eqref#1{equation~\ref{#1}}

\def\1{\bm{1}}

\def\vv{{\bm{v}}}

\def\vz{{\bm{z}}}

\def\mV{{\bm{V}}}

\def\mX{{\bm{X}}}

\DeclareMathAlphabet{\mathsfit}{\encodingdefault}{\sfdefault}{m}{sl}
\SetMathAlphabet{\mathsfit}{bold}{\encodingdefault}{\sfdefault}{bx}{n}

\usepackage{multirow}

\usepackage{xcolor}         
\usepackage{pifont}
\newcommand{\cmark}{\ding{51}}%
\newcommand{\xmark}{\ding{55}}%
\newtheorem{thm}{Theorem}

\FGfinalcopy 

\IEEEoverridecommandlockouts                              
\def\FGPaperID{101} 

\title{\LARGE \bf
Learning Continuous Mesh Representation with \\ Spherical Implicit Surface
}

\author{\parbox{16cm}{\centering
    {\large Zhongpai Gao$^{1,2}$}\\
    {\normalsize
    $^1$ MoE Key Lab of Artificial Intelligence, AI Institute, Shanghai Jiao Tong University, Shanghai, China\\
    $^2$ United Imaging Intelligence, Cambridge MA, USA}}
    \thanks{Copyright notice: 979-8-3503-4544-5/23/\$31.00 ©2023 IEEE. This work was supported by the National Natural Science Foundation of China (61901259) and China Postdoctoral Science Foundation (BX2019208).}
}

\begin{document}

\ifFGfinal
\thispagestyle{empty}
\pagestyle{empty}
\else
\author{Anonymous FG2023 submission\\ Paper ID \FGPaperID \\}
\pagestyle{plain}
\fi
\maketitle

\begin{abstract}

As the most common representation for 3D shapes, mesh is often stored discretely with arrays of vertices and faces. However, 3D shapes in the real world are presented continuously. In this paper, we propose to learn a continuous representation for meshes with fixed topology, a common and practical setting in many faces-, hand-, and body-related applications. First, we split the template into multiple closed manifold genus-0 meshes so that each genus-0 mesh can be parameterized onto the unit sphere. Then we learn spherical implicit surface (SIS), which takes a spherical coordinate and a global feature or a set of local features around the coordinate as inputs, predicting the vertex corresponding to the coordinate as an output. Since the spherical coordinates are continuous, SIS can depict a mesh in an arbitrary resolution. SIS representation builds a bridge between discrete and continuous representation in 3D shapes. Specifically, we train SIS networks in a self-supervised manner for two tasks: a reconstruction task and a super-resolution task. Experiments show that our SIS representation is comparable with state-of-the-art methods that are specifically designed for meshes with a fixed resolution and significantly outperforms methods that work in arbitrary resolutions.

\end{abstract}

\section{INTRODUCTION}

3D shapes in the real world are continuous. While, in the digital world, we usually capture, store, and process 3D shapes in a discrete way. A common representation of 3D shapes is triangulated mesh that structures a 3D shape as arrays of vertices and faces. The precision of mesh representation for 3D shapes is controlled by resolution (i.e., number of vertices). The vertex-based mesh representation has been widely applied in many computer vision and computer graphics applications, e.g., 3D reconstruction \cite{Genova_2018_CVPR, Tran_2018_CVPR, gao2020semi}, shape correspondence \cite{Groueix_2018_ECCV}, virtual avatar \cite{Cao2014}, gesture synthesis \cite{ng2021body2hands}, etc. However, the vertex-based mesh representation is difficult for applications that require various mesh resolutions. In this paper, we propose a continuous representation for meshes. By modeling a mesh as a function defined in a continuous domain, we can process the mesh in an arbitrary resolution as needed.

Closed manifold genus-0 meshes are topologically equivalent to a sphere, hence this is the natural and continuous parameter domain for them, called spherical parameterization \cite{Gotsman2003}. Specifically, spherical conformal parameterization \cite{Baden2018, Choi2020} that preserves the angle and hence the local geometry of the surface is the most important type of parameterization since the angle structure plays an important role in the computation of texture mapping, remeshing, and many other applications. Thus, spherical conformal parameterization provides a one-to-one correspondence between meshes and a sphere such that the spherical coordinate can be considered as the canonical coordinate in a continuous domain for 3D shapes. Inspired by the continuous image representation \cite{chen2020learning} that models an image as an implicit function of the continuous 2D coordinates, we model a mesh as an implicit function of the continuous spherical coordinates. The implicit function can be parameterized by a deep neural network, e.g., multilayer perceptions (MLP) to map each coordinate to the corresponding surface position of the 3D shape. Note that, for a mesh that is not closed manifold genus-0, we always can split the mesh into multiple closed manifold genus-0 meshes with the help of filling holes if necessary.

This paper proposes spherical implicit surface (SIS) for representing mesh in a continuous manner. SIS can represent a mesh with an arbitrary topology. While, in this paper, we mainly focus on the SIS representation for a group of meshes with the same topology, e.g., faces, bodies, and hands. To share knowledge across samples instead of fitting individual implicit function for each mesh, we use an encoder to predict a global feature for each mesh. Then the implicit function is shared by all the meshes while it is conditioned upon the global feature in addition to the spherical coordinates as inputs. At last, the implicit function predicts the 3D position at the given spherical coordinate as the output. Furthermore, instead of using one global feature to encode the whole mesh, we represent a mesh by a set of local features distributed in spatial dimensions (i.e., 3D shape surface). Given a spherical coordinate, the implicit function takes the coordinate information and queries the local features around the coordinate as inputs, then predicts the 3D position at the given coordinate as the output. Either given the global feature or a set of local features of a mesh, the SIS representation can present the mesh in an arbitrary resolution since the spherical coordinates are continuous.

To learn SIS continuous representation from the global feature of a mesh, we train a mesh encoder and an SIS decoder via a reconstruction task in a self-supervised manner. The mesh encoder is built by a convolutional operation named LSA-Conv \cite{GaoAAAI21} to extract the global feature of a mesh. To learn SIS continuous representation from a set of local features of a mesh, we train an SIS encoder, a feature fusion module, and an SIS decoder via a super-resolution task in a self-supervised manner. The SIS encoder takes the vertex information in addition to the spherical coordinate as inputs and predicts the corresponding deep feature as the output. The local feature of a spherical coordinate in a higher resolution is assembled by the feature fusion module which makes use of barycentric coordinates for interpolation. SIS builds a bridge between the discrete and continuous representation in mesh and can naturally exploit the information provided in different resolutions. The SIS representation can present a mesh in an arbitrary resolution, thus it can be trained without resizing ground-truths and achieves better results than methods designed for a certain resolution. We evaluate our approach on the reconstruction and super-resolution task in two 3D shape datasets: human faces (COMA \cite{Ranjan_2018_ECCV}) and human bodies (DFAUST \cite{Bogo_2017_CVPR}).

The contributions of this paper are summarized in below:

1) Taking advantage of that genus-0 meshes are topologically equivalent to a sphere, we use spherical conformal parameterization to map meshes to a sphere as the continuous canonical coordinate for 3D shapes. Then, we introduce a new continuous mesh representation by modeling a mesh as an implicit function of the spherical coordinates, called \textbf{s}pherical \textbf{i}mplicit \textbf{s}urface (SIS).

2) We show how this continuous representation can be used for reconstructing meshes. In addition to the spherical coordinates, the SIS representation either takes the global feature or a set of local features of a mesh as inputs to present the mesh in a continuous manner. For the input of local features, we introduce a feature fusion module that makes use of barycentric coordinates for interpolation to bridge between the discrete and continuous domains.

3) Extensive experiments on COMA \cite{Ranjan_2018_ECCV} and DFAUST \cite{Bogo_2017_CVPR} datasets show that our approach is able to generate high-quality meshes and demonstrate that it compares favorably to state-of-the-art methods designed for discrete domains and outperforms methods designed for continuous domains.

\section{RELATED WORK}

\begin{figure*}[tb!]
    \centering
    \includegraphics[width=1.0\textwidth]{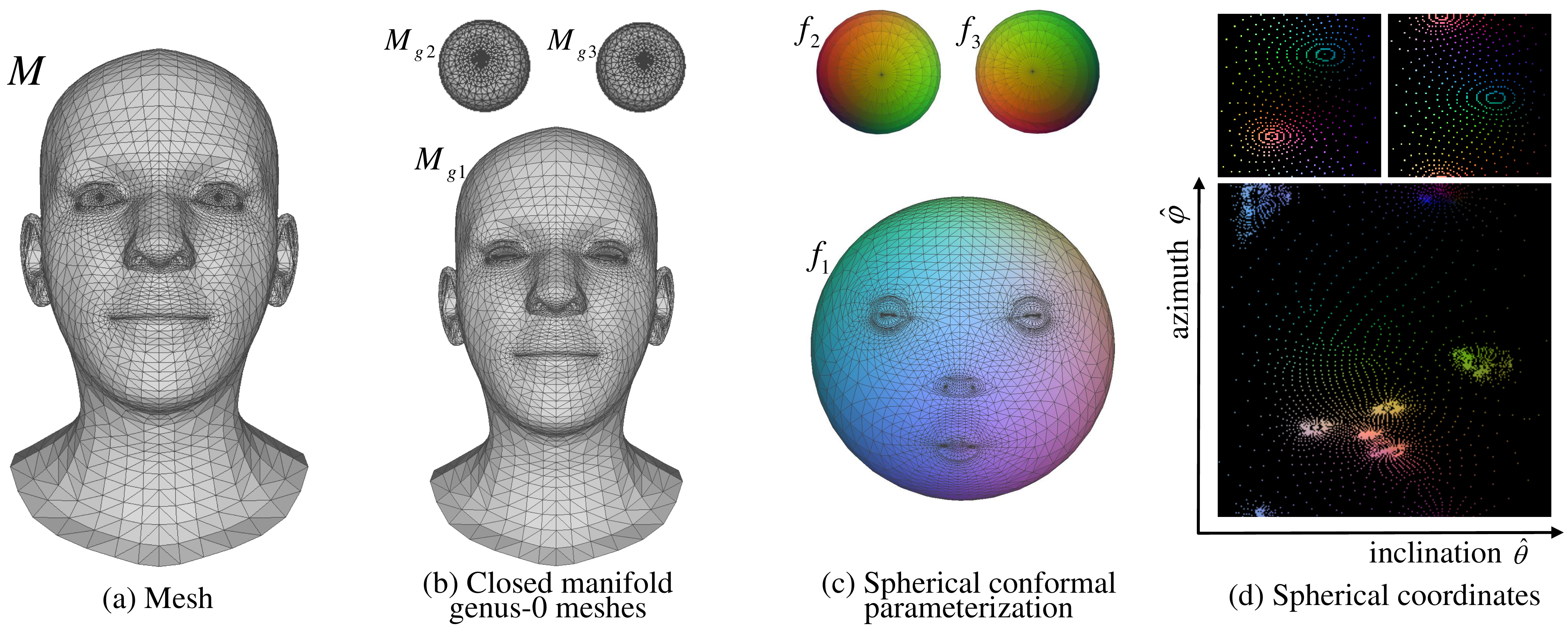}
\caption{Spherical coordinates for a mesh. The facial mesh $M$ in (a) has multiple components and are not a closed manifold genus-0 mesh. It can be split into multiple genus-0 meshes: the left eye $M_{g2}$, the right eye $M_{g3}$, and the rest part $M_{g1}$ in (b). Each genus-0 mesh $M_{gi}$ can be mapped to the unit sphere $S^2$ by applying spherical conformal parameterization $f_i: S^2 \rightarrow M_{gi}$ from the unit sphere $S^2$ to the genus-0 mesh $M_{gi}$. The unit sphere $S^2$ can be parameterized by the normalized inclination angle $\hat{\theta}$ and azimuth angle $\hat{\varphi}$ such that we create a one-to-one correspondence $f_i: (\hat{\theta}, \hat{\varphi})\rightarrow \mV_i$ from the spherical coordinates to the vertices of the genus-0 mesh. The colors in (c) and (d) represent the corresponding vertex position on the mesh.}
\label{fig:spherical}
\end{figure*}

\subsection{Discrete representations for 3D shapes} Discrete representations for learning-based 3D tasks can be mainly categorized as: voxel-based, point-based, and mesh-based. Voxel presentation is a straightforward generalization of pixels to the 3D cases and has been used for discriminative \cite{Daniel2015, Qi_2016_CVPR} and generative \cite{Choy2016, Girdhar2016} 3D tasks. However, voxel representations require memory that grows cubically with resolution. Point clouds \cite{achlioptas2018learning, Fan_2017_CVPR} and meshes \cite{Kanazawa_2018_ECCV, gao2020semi} have been introduced as alternative representations for deep learning. However, point clouds lack the connectivity structure of the 3D shapes and usually require additional post-processing to generate the final 3D shape. Moreover, all the discrete representations are limited to the resolution (i.e., number of points/vertices) that a model can produce. 

In contrast to the discrete representations, our approach leads to continuous surface of 3D shapes. Using deep learning, our approach obtains a more expressive representation that can naturally be integrated into existing 3D shape generation pipelines \cite{gao2020semi, Groueix_2018_ECCV}.

\subsection{Implicit Representations} Implicit representations are continuous and differentiable functions that map coordinates to signal \cite{tancik2020fourfeat}, e.g., images and 3D shapes, and are parameterized as multilayer perceptions (MLP). For images, \cite{chen2020learning} proposed local implicit image function (LIIF) that takes an image coordinate and the 2D deep features around the coordinate as inputs to predict the RGB value at a given coordinate so that the learned representation can present an image in an arbitrary resolution.

For 3D shapes, recent work has investigated implicit representations of continuous 3D shapes that map $xyz$ coordinates to a signed distance function (SDF) \cite{Park_2019_CVPR} or to an occupancy field \cite{Mescheder_2019_CVPR, Peng2020} or to a neural radiance field (NeRF) \cite{Mildenhall2020}. SDF represents a 3D shape's surface by a continuous volumetric field --- the distance of a point to the surface boundary and the sign indicates whether the region is inside or outside of the shape, thus it implicitly encodes a shape's boundary as the zero-level-set of the learned function. Occupancy field is a special case of SDF and only considers the `sign' of SDF values to classify 3D points as inside or outside of a 3D shape. NeRF represents a scene by the volume density and view-dependent emitted radiance of a point and can produce high-fidelity appearance to render photorealistic novel views of complex 3D scenes. Another continuous representation for 3D shapes was introduced by \cite{Groueix_2018_ECCV}, called template deformation (TDeform) that uses an MLP to regress the point-wise deformation of 3D shapes from the template in any resolution.

However, for the implicit representations of SDF \cite{Park_2019_CVPR}, occupancy field \cite{Mescheder_2019_CVPR}, and NeRF \cite{Mildenhall2020}, the coordinates are defined as $xyz$ positions in a volumetric space, which requires large amounts of samples from the volumetric space for training and needs an isosurface extraction algorithm for inference to extract the surface from a trained model. Compared to those implicit representations of 3D shapes, our SIS representation directly works on the surface and is more efficient for both training and inference. Similar to the one-to-one mapping from image coordinates to images, our SIS representation has a one-to-one mapping from the spherical coordinates to the surface of 3D shapes such that we only need to train on the samples of 3D shape vertices and infer a 3D shape simply by inputting the spherical coordinates. 

Even though TDeform \cite{Groueix_2018_ECCV} that defines the coordinates as the template vertices creates a one-to-one mapping from the template to the surface of a 3D shape, the coordinates are $xyz$ positions in a volumetric space and most of the coordinates (except for the template vertices) do not have the corresponding labels, making the network difficult to be trained (not bijective). In contrast, our spherical coordinates are continuous and corresponding to the surface of 3D shapes everywhere (bijective). Thus, our approach is an efficient and effective continuous representation for 3D shapes.  

\section{SPHERICAL IMPLICIT SURFACE}

In this section, we introduce spherical implicit surface (SIS) --- our continuous representation for meshes. First, we apply spherical conformal parameterization to have a one-to-one mapping from a mesh to the unit sphere so that the spherical coordinate can be used as the continuous canonical coordinate for the mesh. Then, we describe how we can learn an SIS network that takes a global feature or a set of local features in addition to the spherical coordinates for 3D shape generations. At last, we introduce the loss function used to train our models.

\subsection{Spherical Coordinate}

\begin{figure*}[tb!]
    \centering
    \includegraphics[width=1\textwidth]{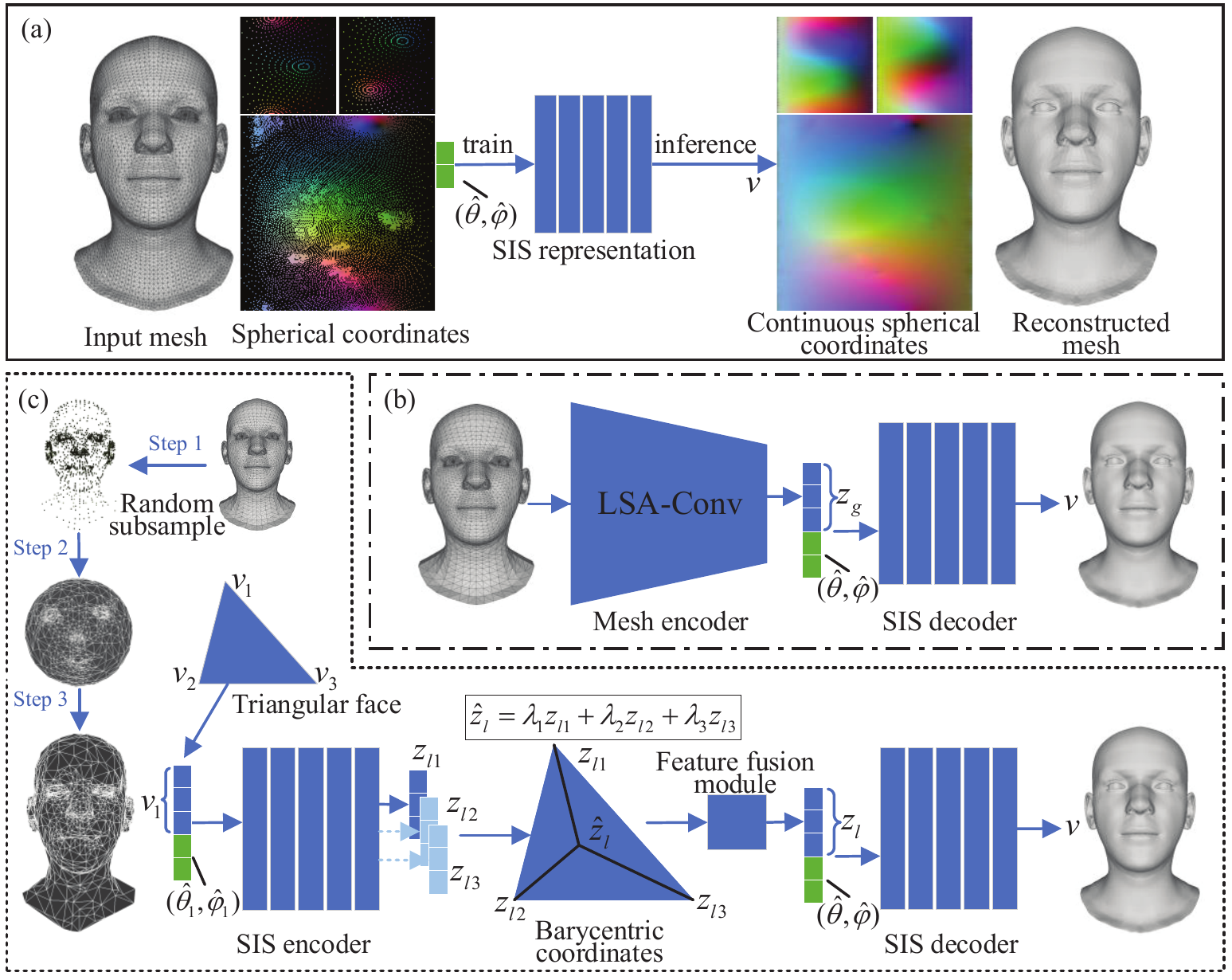}
\caption{Learning to generate spherical implicit surface (SIS) representation for meshes. (a) The SIS representation inputs the spherical coordinates and outputs the mesh vertices and is fitted to an individual mesh. (b) The SIS representation is conditioned with a global feature that is extracted from a mesh using a mesh encoder. (c) The SIS representation is conditioned with a local feature for each spherical coordinate. We input the SIS encoder a subsempled mesh whose topology is created with the help of spherical mapping and output the deep feature for each input vertex. A feature fusion module is introduced to ensemble the local feature for higher-resolution spherical coordinates based on barycentric coordinates.}
\label{fig:architecture}
\end{figure*}

Inspired by images where RGB values and image coordinates (i.e., $xy$ coordinates) are one-to-one corresponding to each other, we seek for a one-to-one mapping between a canonical coordinate and meshes --- a bijectve function.
\begin{thm}\label{as:theorem}
 The `uniformization theorem' guarantees that there is a conformal map $f: S^2 \rightarrow M$ from the unit sphere $S^2$ to any genus-0 mesh $M$, i.e., a smooth, nondegenerate, and globally injective map that preserves both angles and orientation.
 \end{thm}
A mesh can be defined as $M = (\bm{\mathcal{V}}, \bm{\mathcal{F}})$, where $\bm{\mathcal{V}}=\{\vv_1, \ldots, \vv_N\}$ is a set of $N$ vertices and $\bm{\mathcal{F}}\subseteq \bm{\mathcal{V}} \times \bm{\mathcal{V}} \times \bm{\mathcal{V}}$ is a set of triangular faces. The conformal map $f: S^2 \rightarrow M$ can be achieved by applying spherical conformal parameterization \cite{Baden2018,Choi2020} on a genus-0 mesh. As shown in Figure \ref{fig:spherical}, when a mesh is not a genus-0, we can always split the mesh $M$ into multiple submeshes $\{M_1, \cdots, M_K\}$, where $M=\{M_1, \ldots, M_K\}$. We may close the holes of the submeshes $M_i$ as necessary \cite{Attene2010} so that each submesh is a closed manifold genus-0 mesh $M_{gi}$, where $i \in \{1, \ldots, K\}$. The original submesh $M_i$ is a subset of the genus-0 mesh $M_{gi}$, i.e., $M_i=M_{gi}[\mathcal{I}_i]$ where $\mathcal{I}_i$ is the vertex index of $M_i$ in $M_{gi}$. Thus, any mesh can be formulated with one or multiple genus-0 meshes $M=\{M_{g1}[\mathcal{I}_1], \ldots, M_{gK}[\mathcal{I}_K]\}$ where $K\in\mathbb{N}$. Then each genus-0 submesh $M_{gi}$ can find a conformal map $f_i: S^2 \rightarrow M_{gi}$ from the unit sphere $S^2$ to the genus-0 submesh $M_{gi}$. For simplicity and without losing generality, we assume the mesh $M$ is already a closed manifold genus-0 mesh where $K=1$ and $\mathcal{I}=[1, \ldots, N]$.

The surface of the unit sphere $S^2$ can be parameterized by two numbers: its polar (inclination) angle $\theta$ measured from a fixed zenith direction and the azimuthal angle $\varphi$ of its orthogonal projection on a reference plane that passes through the origin and is orthogonal to the zenith, expressed as
\begin{small}
\begin{align}\label{eq:spherical}
\theta &= \arctan\frac{\sqrt{x^2+y^2}}{z} \in [0, \pi], & \varphi &= \arctan\frac{y}{x}\in [-\pi, \pi],\\
\hat{\theta} &= \frac{\theta}{\pi}\in [0, 1],& \hat{\varphi} &=\frac{\varphi}{2\pi} + 0.5\in [0, 1],
\end{align}
\end{small} where $\hat{\theta}$ and $\hat{\varphi}$ are the normalized inclination and azimuthal angle of a point on the unit sphere. Thus, we create a one-to-one correspondence $f: (\hat{\theta}, \hat{\varphi})\rightarrow \mV$ from the spherical coordinates to the mesh vertices, i.e., $\vv_{j} = f(\hat{\theta}_j, \hat{\varphi}_j)$, where $j\in\{1, \ldots, N\}$. As shown in Fig. \ref{fig:architecture}a, the implicit function $f$ that is parameterized as an MLP network can be trained in a supervised manner. The trained implicit function $f$ is a continuous representation for the mesh, called spherical implicit surface (SIS) representation. During the inference phase, we can take $(\hat{\theta}, \hat{\varphi})$ continuously to generate the 3D shape in a higher resolution.

Though an MLP networks are universal function approximations \cite{HORNIK1989359}, directly inputting the coordinates $(\hat{\theta}, \hat{\varphi})$ performs poorly at representing high-frequency variation in geometry and Fourier feature mapping enables an MLP network to learn high-frequency functions \cite{tancik2020fourfeat}. Inspired by NeRF \cite{Mildenhall2020}, we encode the spherical coordinates as
\begin{equation}\label{eq:fourier}
\resizebox{.895\hsize}{!}{
$\xi(p) = (\sin(2^0\pi p), \cos(2^0\pi p), \ldots, \sin(2^{L-1}\pi p), \cos(2^{L-1}\pi p))$},
\end{equation}
where $p=(\hat{\theta}, \hat{\varphi})$ and $L=10$ in our experiments. Though Fourier feature mapping $\xi(\cdot)$ has been used in NeRF, applying it on our spherical coordinates is physically more meaningful than on the $xyz$ coordinates used in NeRF since the spherical coordinates $(\hat{\theta}, \hat{\varphi})$ are defined in angles as presented in Eq. (\ref{eq:spherical}) and Fig. \ref{fig:spherical}d are periodic, which is naturally suitable for Fourier feature mapping.

\subsection{Condition with Global Feature}

Instead of fitting the implicit function $f$ to an individual mesh $M$, we propose an SIS representation that is shared by a group of meshes, which can be achieved by conditioning an observation of that mesh on the input in addition to the spherical coordinates. We train the model in a self-supervised manner via a reconstruction task. The observation of a mesh can be considered as a global feature $\vz_g$ extracted by a mesh encoder, as shown in Fig. \ref{fig:architecture}b. During the inference phase, we can use the implicit function to reconstruct a mesh in an arbitrary resolution given its global feature. Thus, the implicit function (i.e., SIS decoder) can be expressed as
\begin{align}\label{eq:global}
\vv=f(\vz_g, \hat{\theta}, \hat{\varphi}),
\end{align}
where $\vz_g=en_g(M)$ and $en_g(\cdot)$ is the mesh encoder built by convolutional operations and LSA-Conv \cite{GaoAAAI21} is used in our experiments.

\subsection{Condition with Local Feature}

To make the SIS representation more expressive, instead of using one global feature to encode the whole mesh, we encode a mesh by a set of local features distributed in spatial dimensions such that each of them stores information about its local area. We train the model in a self-supervised manner via a super-resolution task. Thus, the input is noisy sparse point cloud that is randomly sampled from the mesh (step 1 in Fig. \ref{fig:architecture}c). Based on the spherical mapping, we can find the points on the sphere corresponding to the point cloud (step 2 in Fig. \ref{fig:architecture}c). Then, we can easily and consistently build a topology connection for the corresponding points on the sphere, which is the same for the point cloud, thus we build a subsampled mesh (i.e., a lower resolution mesh) for the randomly sampled point cloud (step 3 in Fig. \ref{fig:architecture}c).

For a subsampled mesh, the SIS encoder maps each vertex $\vv_i$ to a deep feature $\vz_{li}$. Note that, the spherical coordinates of the subsampled mesh can correspond to any point on the sphere since the SIS encoder is a continuous representation. The SIS decoder is also a continuous representation and may take spherical coordinates that are not provided in the subsampled mesh, i.e., spherical coordinates in a higher resolution. Thus, the SIS encoder cannot provide the deep feature for those higher-resolution spherical coordinates. We propose a feature fusion module based on barycentric coordinates to obtain the local feature given any spherical coordinate.

Given a pair of spherical coordinate $(\hat{\theta}, \hat{\varphi})$, we first find the triangular face that contains the spherical coordinate on the sphere that has the same topology of the subsampled mesh. We denote the spherical coordinates of the triangular vertices on the sphere as $[(\hat{\theta}_1, \hat{\varphi}_1), (\hat{\theta}_2, \hat{\varphi}_2)$, $(\hat{\theta}_3, \hat{\varphi}_3)]$ and denote the triangular vertices of the subsampled mesh as [$\vv_1, \vv_2$, $\vv_3]$. The deep features of the triangular vertices are $\vz_{l1}=en_l(\vv_1), \vz_{l2}=en_l(\vv_2)$, and $\vz_{l3}=en_l(\vv_3)$, where $en_l(\cdot)$ is the SIS encoder. We can calculate the barycentric coordinates for the spherical coordinate $(\hat{\theta}, \hat{\varphi})$ relative to the three triangular vertices as $[\lambda_1, \lambda_2, \lambda_3]$ where $\sum_{i=1}^{3}\lambda_i=1$. Thus, based on the barycentric coordinates, we can obtain a coarse deep feature for the spherical coordinate $(\hat{\theta}, \hat{\varphi})$ as,
\begin{align}\label{eq:barycentric}
\hat{\vz}_l=\lambda_1\vz_{l1} + \lambda_2\vz_{l2} + \lambda_3\vz_{l3}.
\end{align}
The feature fusion module ensembles the local feature for the spherical coordinate $(\hat{\theta}, \hat{\varphi})$ as
\begin{align}\label{eq:fusion}
\vz_l=\hat{\vz}_l \oplus(\hat{\vz}_l-\vz_{l1})\oplus(\hat{\vz}_l-\vz_{l2})\oplus(\hat{\vz}_l-\vz_{l3})\oplus \mathbf{\lambda}.
\end{align}
At last, the implicit function (i.e., SIS decoder) can be expressed as
\begin{align}\label{eq:local}
\vv=f(\vz_l, \hat{\theta}, \hat{\varphi}).
\end{align}

\subsection{Loss Function Design}
Our SIS representation defines the coordinates that are one-to-one corresponding to the surface of 3D shapes. Thus, we can train the models in a self-supervised manner for each vertex of 3D shapes. First, the L1 reconstruction loss of vertices is used as
\begin{align}\label{eq:loss1}
L_{rec} = \left\|\mV-\hat{\mV}\right\|_1,
\end{align}
where $\mV$ is the ground truth vertices and $\hat{\mV}$ is the vertices predicted by our SIS decoders. Then, Laplacian regularization is introduced to help the mesh reconstruction. Laplacian term is defined as the difference between the vertex and the mean of its one-ring neighbors, expressed as $\mV_i-\frac{1}{|\mathcal{N}_i|}\sum_{j\in\mathcal{N}_i}\mV_j$ where $\mV_i$ is the $i$th vertex and $\mathcal{M}_i$ is the indices of its one-ring neighbors of $\mX_i$. We propose a Laplacian loss that calculates the Laplacian term difference between the ground truth vertices and the predicted vertices, expressed as
\begin{tiny}
\begin{align}\label{eq:loss2}
L_{lap} = \sum_{i\in\mathcal{M}}\left\|\left(\mV_i-\frac{1}{|\mathcal{N}_i|}\sum_{j\in\mathcal{N}_i}\mV_j\right) - \left(\hat{\mV}_i-\frac{1}{|\mathcal{N}_i|}\sum_{j\in\mathcal{N}_i}\hat{\mV}_j\right)\right\|_1,
\end{align}
\end{tiny}where $\mathcal{M}$ is the vertex indices of the mesh. The overall loss function is defined as
\begin{align}\label{eq:loss}
L=L_{rec} + \gamma L_{lap},
\end{align}
where $\gamma=0.05$ in our experiments. During the inference phase, we can output a 3D shape simply by inputting the spherical coordinates with a global feature or a local feature to the SIS decoders.

\section{EXPERIMENTS AND EVALUATION}
\label{sec:exp}
In this section, we evaluate our SIS representation on two different 3D shape datasets in two tasks: reconstruction task and super-resolution task. For the reconstruction task, we input meshes with fixed topology and condition the SIS representation with a global feature. For the super-resolution task, we input point clouds that are randomly downsampled from meshes and condition the SIS representation with a local feature that is assembled by a feature fusion module based on  baryccentric coordinates.

\paragraph{Datasets} In line with \cite{GaoAAAI21}, we evaluate our model on two datasets: COMA \cite{Ranjan_2018_ECCV} and DFAUST \cite{Bogo_2017_CVPR}. COMA is a human facial dataset that consists of 12 classes of extreme expressions from 12 different subjects. The dataset contains 20,466 3D meshes that were registered to a common reference template with 5,023 vertices. DFAUST is a human body dataset that collects over 40,000 real meshes, capturing 129 dynamic performances from 10 subjects. The meshes were also registered to a common reference topology that has 6,890 vertices. Both two datasets are split into training and test set with a ratio of 9:1 and randomly select 100 samples from the training set for validation. The test samples are obtained by picking consecutive frames of length 10 uniformly at random across the sequences. All of the 3D meshes are standardized to have a mean of 0 and standard deviation of 1 to speed up the training.

\paragraph{Training} We use Adam \cite{kingma2014adam} optimizer with learning rate 0.001 and reduce the learning rate with decay rate 0.98 in every epoch. The batch size is 64 and total epoch number is 200. Weight decay regularization is used for the network parameters. We implemented the models in PyTorch and trained on the same machine with an AMD 3700X @3.6GHz CPU and an NVIDIA RTX2080Ti GPU.

\paragraph{Architecture} As shown in Fig. \ref{fig:architecture}b, we adopt the mesh encoder from \cite{GaoAAAI21}. The encoder has four LSA-Conv layers with downsampling. The conv layers have channel sizes of [3, 16, 32, 64, 128] and meshes are downsampled with ratios of [4, 4, 4, 4]. A fully connected layer outputs the latent vector of 64 dimension that represents the 3D mesh.

For COMA dataset, as shown in Fig. \ref{fig:spherical}, the template facial mesh is split into three genus-0 meshes: left eye, right eye, and the rest part. Thus, we need three SIS networks to represent the facial meshes. For DFAUST dataset, the template body mesh is split into six genus-0 meshes: head, torso, left arm, right arm, left leg, and right leg. Thus, we need six SIS networks to represent the body meshes. Each SIS network is an MLP with a skip connection in the middle layer. As shown in Fig. \ref{fig:architecture}, the SIS encoders are conditioned with vertices in addition to the spherical coordinates and output the corresponding deep features. The SIS decoders are conditioned with local features in addition to the spherical coordinates and output the vertices of 3D shapes.

\subsection{Task 1: Reconstruction}

\begin{table}[t]
\centering
\caption{Comparison of reconstruction errors for the models of LSA-small \cite{GaoAAAI21}, FeaStNet \cite{Verma_2018_CVPR}, and template deformation (TDeform) \cite{Groueix_2018_ECCV} when latent size $d=64$. For a fair comparison, we adjust the channel sizes to have around the same parameter size. \cmark ~ represents the decoder can infer 3D shapes in an arbitrary resolution. \xmark ~ represents the decoder can only infer 3D shapes in a fixed resolution of the template. The `time (s)' denotes the duration to infer the test sets.}
\label{tb:reconstruction}
\resizebox{0.48\textwidth}{!}{
\begin{tabular}{l|l|c|ccc}
\toprule
 &&& L2(mm)$\downarrow$ & time (s)$\downarrow$ & parm \# \\
\midrule
  \parbox[t]{2mm}{\multirow{4}{*}{\rotatebox[origin=c]{90}{DFAUST}}} &LSA-small~\cite{GaoAAAI21}&\xmark&\textcolor{red}{3.679}&\textcolor{blue}{3.992}& 547K\\  
  &FeaStNet~\cite{Verma_2018_CVPR}&\xmark & \textcolor{blue}{3.769}&5.146& 548K \\
  &TDeform~\cite{Groueix_2018_ECCV} &\cmark & 6.897 &4.391& 549K \\
  &\bf SIS (ours)& \cmark & 4.737 &\textcolor{red}{3.273}& 547K \\
 \midrule
 \parbox[t]{2mm}{\multirow{4}{*}{\rotatebox[origin=c]{90}{COMA}}} & LSA-small~\cite{GaoAAAI21}&\xmark&
\textcolor{red}{0.172}&\textcolor{blue}{5.615}& 378K \\
 &FeaStNet~\cite{Verma_2018_CVPR}&\xmark & 0.208 &9.969& 378K\\
 &TDeform~\cite{Groueix_2018_ECCV} &\cmark & 0.946 &5.434& 378K\\
 &\bf SIS (ours)& \cmark & \textcolor{blue}{0.179} &\textcolor{red}{5.357}& 378K\\
 
\bottomrule
\end{tabular}}
\end{table}

For the reconstruction task as shown in Fig. \ref{fig:architecture}b, we compare three existing methods: LSA-Conv, FeaStNet, and template deformation (TDeform) when the latent space is 64. TDeform proposed by \cite{Groueix_2018_ECCV} uses the template as the canonical coordinate of meshes. Similar to SIS representation, the TDeform decoder is also built by an MLP network that predicts the deformation of the vertices of a mesh relative to the template vertices. During the inference phase, we can provide a higher-resolution template to predict 3D shapes that have the same resolution as the template. Table \ref{tb:reconstruction} shows the quantitative results. For a fair comparison, we adjust the channel sizes for each methods to have around the same model size. For methods that can infer 3D shapes in an arbitrary resolution (labeled as \cmark in Table \ref{tb:reconstruction}), our SIS representation outperforms TDeform in both DFAUST and COMA datasets. For COMA dataset, our representation even performs better than FeaStNet that only works in a fixed resolution. In terms of time comlexity, the proposed SIS is the most time-efficient compared with other methods since SIS networks are simply MLPs.

Note that, we split both the facial template and body template into multiple genus-0 submeshes and each submesh requires an SIS network. In order to control the overall model size to be around the same with other methods, the parameter size for each SIS network is small. As shown in Table \ref{tb:reconstruction}, we split more parts for the body template than for the facial template, thus, each body part has a smaller SIS network and only has 5 or 6 layers with 131 channel size, resulting in larger errors in DFAUST dataset than COMA dataset compared to other methods. However, even we need an extra SIS network for the eyes in COMA dataset, our SIS representation is marginally on par (0.179 vs. 0.172) with LSA-small that is the current best convolutional operation designed for meshes. Even though controlling the overall model size to be the same with other methods is not favorable for our setting, our SIS representation consistently outperforms TDeform that uses one but deeper and larger MLP network. For TDeform, the input could be any $xyz$ point in the volumetric space while only the points of template vertices are trained with labels. Thus, most of samples (except for the template vertices) are not trained for the implicit function of TDeform, i.e., undersampling occurs.

\subsection{Task 2: Super-resolution}

\begin{figure}[tb!]
    \centering
    \includegraphics[width=0.485\textwidth]{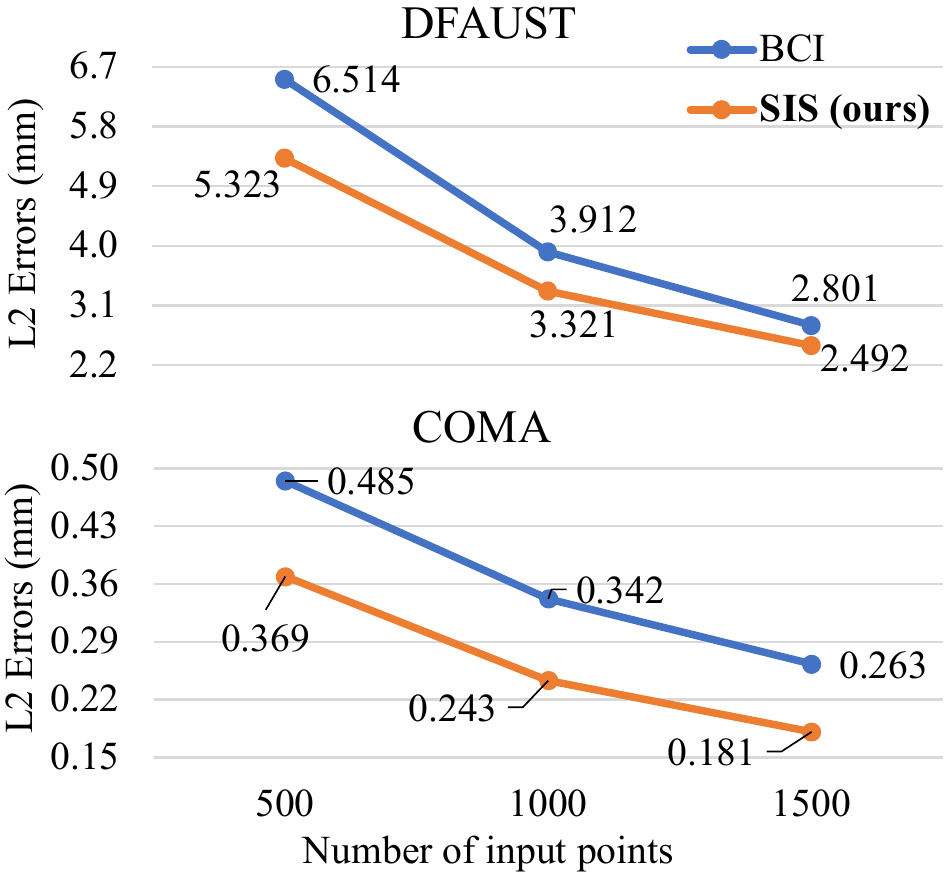}
\caption{Comparison of reconstruction errors between our SIS representation and BCI (barycentric interpolation) for the super-resolution task. We train the models with 1,000 input points and infer the models with input points of 500, 1,000, and 1,500.}
\label{fig:results_super}
\end{figure}


\begin{figure*}[tb!]
    \centering
    \includegraphics[width=0.99\textwidth]{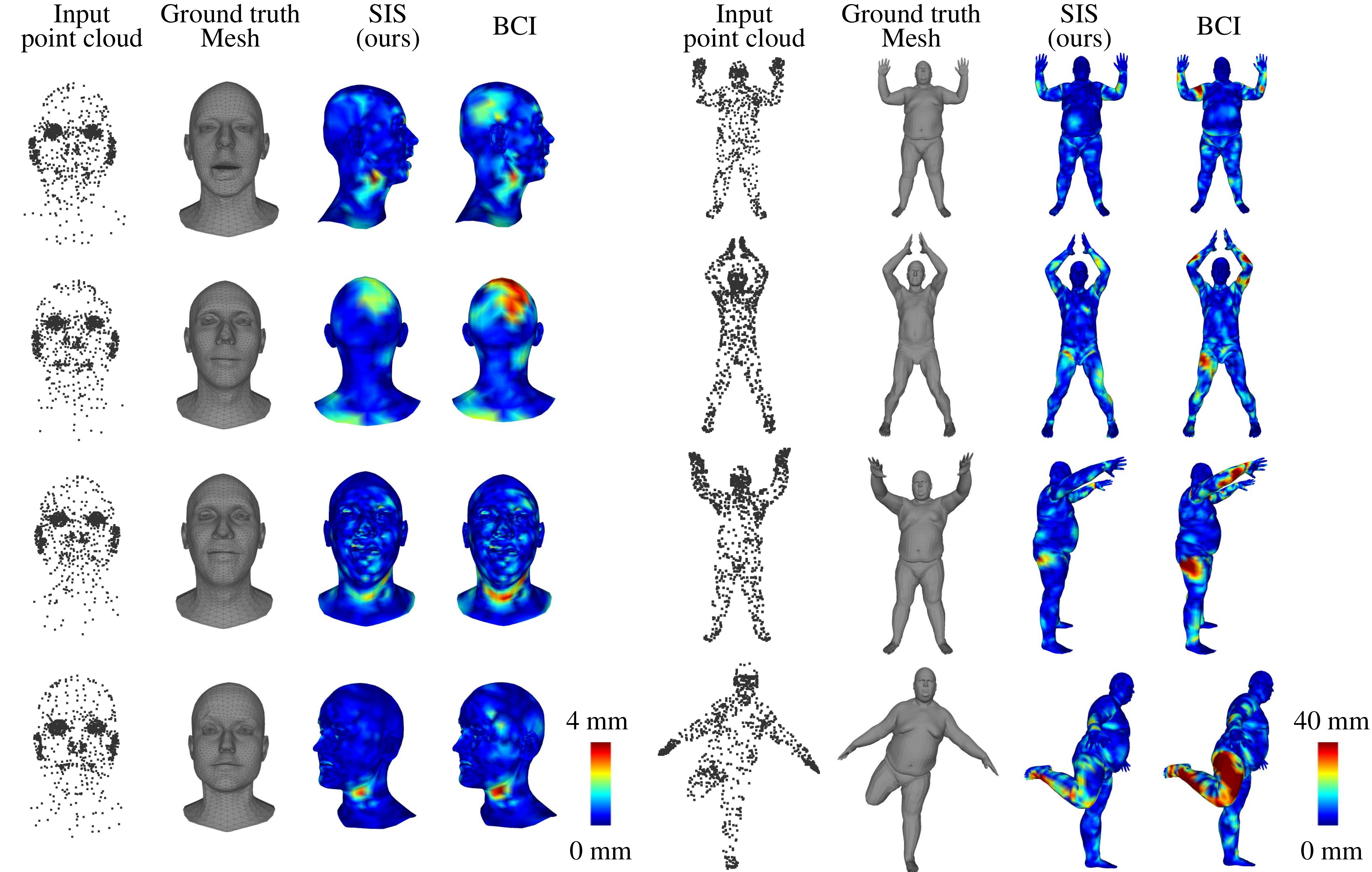}`
\caption{Qualitative results of the super-resolution task. The per-vertex Euclidean errors produced by our SIS representation and BCI are visualized in colormap. The input point cloud has 1,000 points that are randomly sampled from the ground truth mesh. The left and right are some examples wihh varisou facial expressions and body poses from the test sets of COMA and DFAUST datasets.}
\label{fig:results}
\end{figure*}

For the super-resolution task as shown in Fig. \ref{fig:architecture}c, we randomly sample 1,000 points from a mesh as the input to train our models in a self-supervised manner in DFAUST and COMA datasets. We compare our method with a traditional algorithm: barycentric interpolation (BCI). BCI interpolates the vertex of a given spherical coordinate based on the barycentric coordinates that are calculated from the triangular face on the sphere. For instance, when the triangular vertices are $[\vv_1, \vv_2, \vv_3]$ and the barycentric coordinates are $[\lambda_1, \lambda_2, \lambda_3]$ where $\sum_{i=1}^{3}\lambda_i=1$, the interpolated vertex is expressed as $\vv = \lambda_1\vv_1 + \lambda_2\vv_2 + \lambda_3\vv_3$. 

We evaluate our approach and BCI with three different numbers of input points: 500, 1,000, and 1,500. As shown in Fig. \ref{fig:results_super}, our SIS representation consistently outperforms BCI in both DFAUST and COMA datasets for all the different numbers of input points, which demonstrates the robustness of our SIS representation. The qualitative results presented in Fig. \ref{fig:results} also show that our approach produces smaller errors than BCI for both DFAUST and COMA datasets in various body poses and facial expressions.

\subsection{Ablation Study}

\begin{figure}[tb!]
    \centering
    \includegraphics[width=0.488\textwidth]{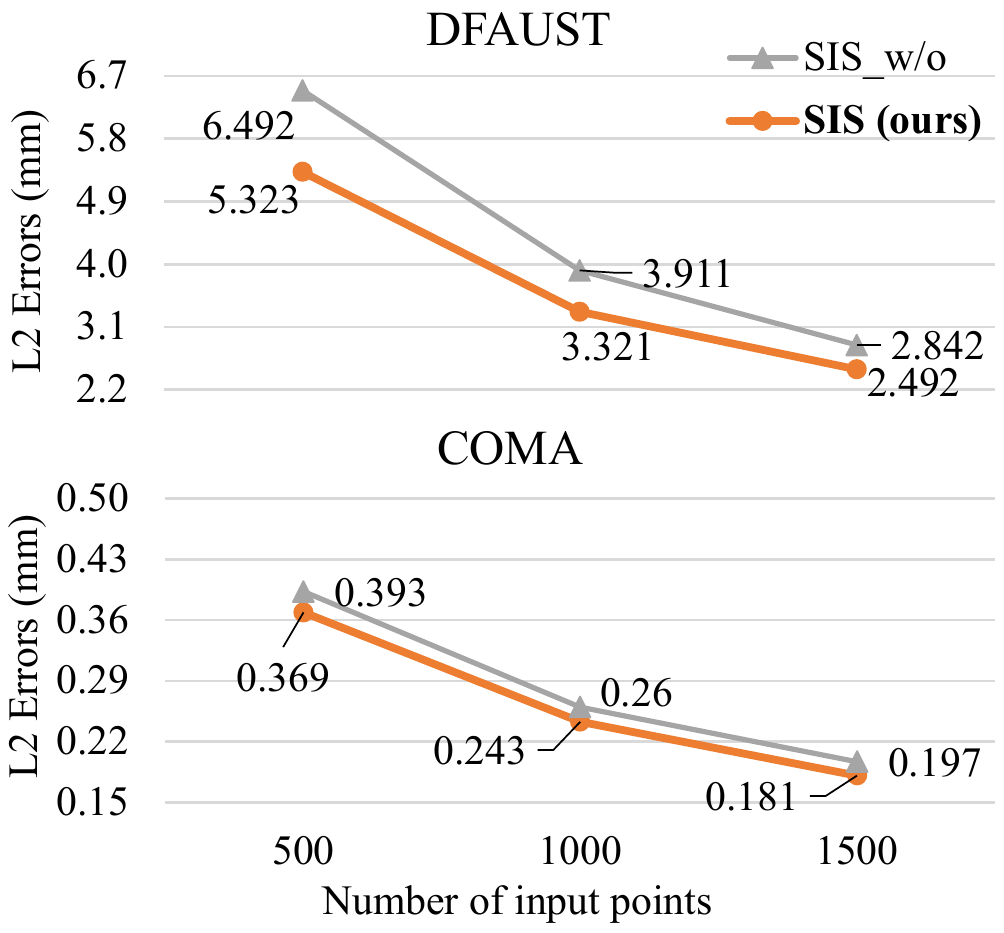}
\caption{Ablation study for of the feature fusion module in the super-resolution task. ``SIS\_w/o'' means we use the coarse feature in Eq. \ref{eq:fusion} as the input of the SIS decoder without the feature fusion module. We train the models with 1,000 input points and infer the models with input points of 500, 1,000, and 1,500.}
\label{fig:ablation}
\end{figure}

For the super-resolution task, we design a feature fusion module to ensemble the deep features for the local feature of a given spherical coordinate. To evaluate the effectiveness of the feature fusion module, we conduct an ablation study where we simply use the coarse deep feature $\hat{\vz}_l$ (Eq. 5) as the local feature of a given spherical coordinate without the feature fusion module, denoted as \emph{SIS\_w/o} in Figure \ref{fig:ablation}. For both the COMA and DFAUST datasets, our SIS representation with the feature fusion module outperforms \emph{SIS\_w/o} consistently with different input points. This is berceuse the feature fusion module considers the edges between the coarse deep feature with the deep features of the triangular vertices and provides more local structure around the spherical coordinate. Thus, our SIS representation with the feature fusion module can generate 3D shapes with more details.

\section{CONCLUSION AND DISCUSSIONS}
\label{sec:discussion}
\subsection{Conclusion}
We propose to learn the continuous representation for meshes, which is fulfilled by our devised spherical implicit surface (SIS) technique. SIS builds a bridge between the discrete and continuous representation in mesh and can naturally exploit the information provided in different resolutions. To share knowledge across samples, we condition the SIS representation with a global feature or a set of local features of a mesh. We show that this continuous representation technique can be effectively applied for downstream tasks like reconstruction and super-resolution of 3D shapes.

\subsection{Limitations} The SIS representation for meshes is similar to the implicit function for images \cite{tancik2020fourfeat}. When the resolution of a mesh is too low, the SIS representation may overfit to the small amount of training samples and cannot generalize well to the whole surface of the mesh. Thus, high-resolution meshes are more favorable to train an SIS network. Furthermore, the experimented datasets may not fully reflect the challenges in real-world scenarios.

\subsection{Future works} 

In this work, we split a mesh template into multiple genus-0 submeshes and train an independent SIS network for each submesh. In the future, we can create a shared SIS network for all the submeshes to reduce the model size. Furthermore, currently, we simply encode the spherical coordinates with Fourier feature mapping. More advance coordinate encoding methods \cite{mueller2022instant} can be integrated to our SIS representation. 

{\small
\bibliographystyle{ieee}
\bibliography{mesh21}
}

\end{document}